# Deep Learning Based Classification System For Recognizing Local Spinach

Mirajul Islam[1], Nushrat Jahan Ria[1], Jannatul Ferdous Ani[1], Abu Kaisar Mohammad Masum[1], Sheikh Abujar[2], Syed Akhter Hossain[3]

[1] Department of Computer Science and Engineering, Daffodil International University, Dhaka 1209, Bangladesh
[2] Department of Computer Science and Engineering, Independent University Bangladesh, Dhaka 1229, Bangladesh
[3] Department of Computer Science and Engineering, University of Liberal Arts Bangladesh, Dhaka 1207, Bangladesh

```
{merajul15-9627, nushrat15-9771, jannatul15-10483,
         mohammad15-6759}@diu.edu.bd
           sheikh.csesets@iub.edu.bd
            akhter.hossain@ulab.edu.bd
```

**Abstract.** A deep learning model gives an incredible result for image processing by studying from the trained dataset. Spinach is a leaf vegetable that contains vitamins and nutrients. In our research, a Deep learning method has been used that can automatically identify spinach and this method has a dataset of a total of five species of spinach that contains 3785 images. Four Convolutional Neural Network (CNN) models were used to classify our spinach. These models give more accurate results for image classification. Before applying these models there is some preprocessing of the image data. For the preprocessing of data, some methods need to happen. Those are RGB conversion, filtering, resize & rescaling, and categorization. After applying these methods image data are preprocessed and ready to be used in the classifier algorithms. The accuracy of these classifiers is in between 98.68% - 99.79%. Among those models, VGG16 achieved the highest accuracy of 99.79%.

**Keywords:** Spinach Recognition, Convolutional Neural Network (CNN), Deep Learning, Image Classification, Evaluation metric.

## 1      Introduction

Spinach is one of the favorable foods for the human body. It strengthens the bones of our body. It produces vitamin-A, vitamin-C, vitamin-E, vitamin-K, potassium, magnesium, iron, calcium, copper, phosphorous, zinc, selenium, folate, betaine, folic acid, protein, niacin, omega-3 fatty acids, and many more important nutrient [1]. It greatly enhances our ability to repel diseases. Also, some species are very important for our hair, eyes, heart, kidney, and skin. Spinach is a very popular food in the rural areas of Bangladesh. There are 107 types of spinach available in Bangladesh [2]. Although many of these are available in the market of rural areas, only a few species are found in urban areas. The spinach looks quite similar so most of the people & children of the new generation usually cannot get to recognize them. Even they do not know the name of these spinach. As a solution, we used several Convolution Neural Networks(CNN) to automatically classify different species of spinach based on spinach's leaves. We used five species of spinach for this research work. These are Jute spinach (Corchorus olitorius) which is locally known as Pat Shak, Malabar spinach (Basella alba) which is locally known as Pui Shak, Red spinach (Amaranthus gangeticus) which is locally known as Lal Shak, Taro spinach (Colocasia esculenta) which is locally known as Kochu Shak and Water spinach (Ipomoea aquatica) which is locally called as Kolmi Shak [2] . We preprocessed the data then we trained four different models for this classification. We used InceptionV3, Xception, VGG19, and VGG16 to identify those spinach.

CNN based image classification performs a vast advancement. There are two types of image classification, one is supervised and another one is unsupervised. Deep neural networks play the most important role in image analysis. Image classification is the primary domain, in which deep neural networks are being used. The model takes the given input images and provides classified output images for identifying whether the image represents a specific class or not. CNN model that performs perfect image classification accuracy for different images. Inception V3 architecture is one of the best models for image data analysis. Human performance with efficiency has been accomplished by Inception V3. The CNN architecture uses three-dimensional convolutions to read different input images. CNN's are the recent state-of-the-art methods in image classification. Natural images are learned by the CNNs, showing strong performance and encountering the accuracy of human expert systems. Finally, these statements conclude that CNNs can classify different images into different classes more accurately.

The remainder of this paper is organized as follows: Section 2 describes some of the work that is similar to our work. Section 3 describes our dataset, pre-processing steps, and the entire process. In section 4 shows the results and graphs for each model and compares each of them. And in section 5 describes our work summary and future work.

## 2      Related work

In this sector of leaf recognition and classification, we studied some related research work. Stephen Gang Wu et al. [3] proposed a Probabilistic Neural Network (PNN) with data processing techniques and images. At first, they captured digital leaf images. Then they processed those images. They extracted the features from the images. The features are orthogonalized by Principal Components Analysis (PCA). After that, they trained the data in Probabilistic Neural Network (PNN) & then tested the data in Probabilistic Neural Network (PNN). Finally, they display & compare the results. Their proposed algorithm shows 90% accuracy for the model. Guan Wang et al. [4] proposed a deep learning method to automatically and accurately estimate disease severity security, disease management, and yield loss prediction for diagnosing plant disease severity. This method used an image-based plant disease recognition, this work proposes deep learning models for image-based automatic diagnosis of plant disease severity. They used the VGG16 model for this method because this model gave a better performance than other models. VGG16 model achieves an accuracy of 90.4%. Yu Sun et al. [5] proposed a recognition plant model by using the BJFU100 dataset and deep learning model. This model contains 10,000 images of 100 ornamental plant species. BJFU100 has collected data from natural scenes by mobile devices. The Deep Residual Network method has been used to automatically identify the required representations of the classification. There is a 26 layer deep learning model consisting of 8 residual building blocks designed for uncontrolled plant identification. The proposed model recognition accuracy is 91.78% on the BJFU100 dataset. In recent years, Susu Zhu et al. [6] proposed a system which is the possibility of the freshness identification of Spinach preserved at different temperatures by using Hyperspectral Imaging. They have applied three models. But there are ELM models performed best and the accuracy was achieved 100% of ELM models for the two spectral systems in the freshness detection of spinach leaves preserved at 4◦C and 20◦C. In this paper, Hulya Yalcin et al. [7] are mainly using a Convolutional Neural Network(CNN) model for classifying different plant species and identifying different plant species. Their proposed CNN architecture can automatically classify images of sixteen kinds of plants and achieve an accuracy of 97.47%. Aydin Kayaa et al. [8] in plant classification models (2019), machine learning algorithms Deep Neural Networks (DNNs) have been applied to different data sets. They had designed and implemented five classification models including the baseline model (end-to-end CNN model) by applying four transfer learning strategies on deep learning-based models. Arun Priya C. et al.[9] implemented plant leaf recognition through the kNN method. Firstly, Classifier tested with a Flavia dataset and a real dataset then compared with the KNN approach of plant leaf recognition. KNN approach had produced high accuracy and less execution time.

## 3  Research Methodology

In our work at first, we prepare our dataset with raw images, then we have divided those images into two portions, one is the training set and another is the test set, shown in Table 1. The five types of spinach are divided separately in both training and test sets. Then The images from both sets are prepared for input into the model through pre-processing, shown in Fig. 2. Then four CNN models were trained with the data from our training set to classify the spinach. We used four CNN models to classify our data and evaluate their performances. The accuracy of the model has been measured on the ability to classify the data in the test set. These four models are InceptionV3[10][11], Xception[12], VGG19[13], and VGG16[13].

### 3.1  Collection & Properties of Dataset

We have collected a total of 3,785 images of five types of fresh spinach leaves. These images have been collected from the local vegetable market and spinach fields. All images are captured in a fixed background. The images have been divided into two parts. The training set contains 80% of the images and the test set which contains 20% of images for testing the model, shown in Table 1. From the data distribution, it is a stabilized dataset that we can convey.

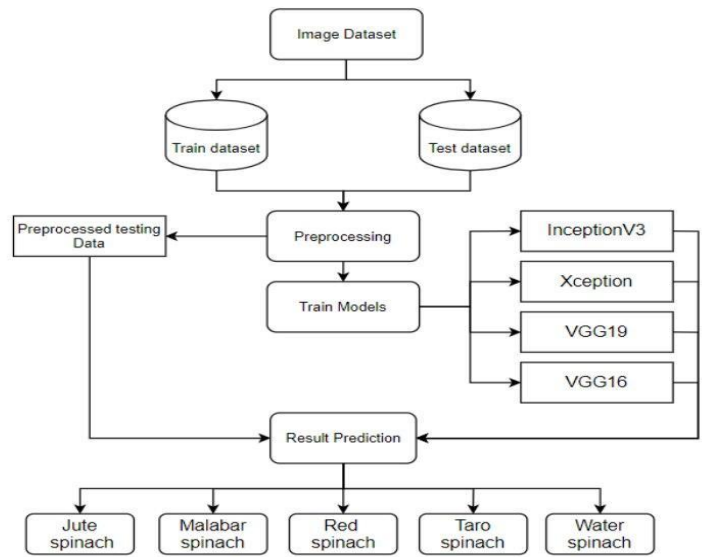

**Fig. 1.** Classification procedure.

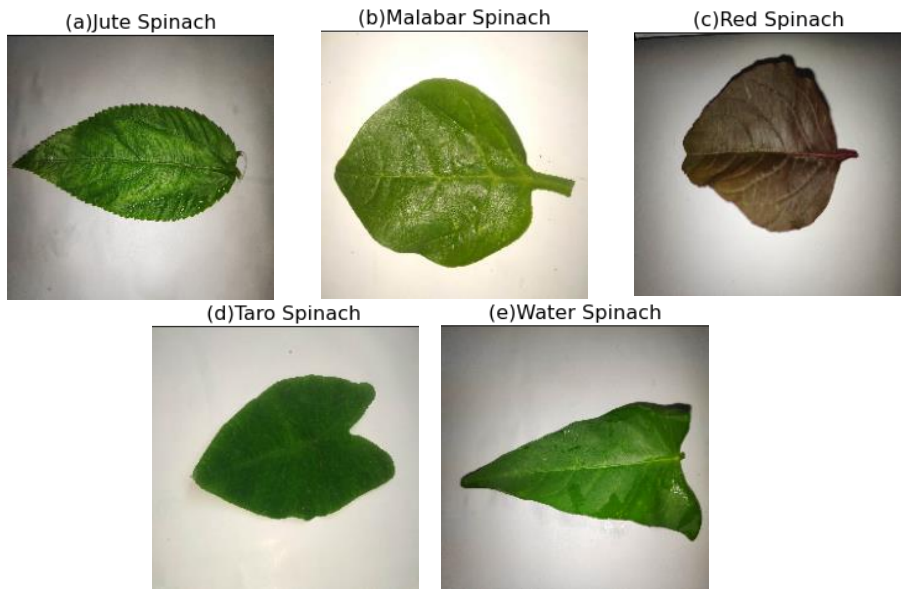

**Fig. 2.** Five types of spinach. (a)Jute Spinach, (b)Malabar Spinach, (c)Red Spinach, (d)Taro Spinach, (e)Water Spinach.

### 3.2 Data Pre-processing

Data preprocessing is a common method to process the data for implementing the model. In pre-processing, we take the raw images then apply different techniques to prepare the input data for the next stage. The main purpose of this image pre-processing is to improve the quality of the information of the data so that the machine can understand and analyze those data easily. In this data Pre-processing process, the images have been converted into RGB images. Then we filter or decrease the noise of the raw images. After that resize & rescale those data to process and then categorize the input data. Then the data is preprocessed & ready to train in the model. After preparing our dataset, to input, all the images into the model all images have been resized into height 224, width 224, and RGB color channel 3. Because our dataset contains different sizes of the images. 224x224x3 is the input shape of our model.

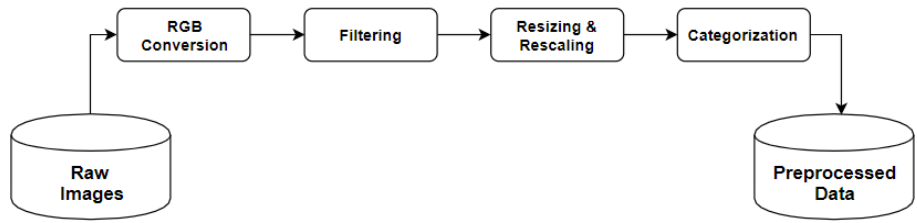

**Fig. 3.** Pre-processing method.

Then performing an image data generator, which is the main part of image preprocessing. At first, all the images from the training and test set have been rescaled the values between 0 and 1.This process is known as Min-Max Normalization(MMN) [14]. Usually, the value of an RGB image is between 0 and 255. The equation for MMN as follows,

$$x(scaled) = \frac{x - min(x)}{max(x) - min(x)} (nMax - nMin) + nMin \qquad (1)$$

Where nMax, nMin is the maximum and minimum value.

## 4 Experiment & Output

**Table 1.** Distribution data into Training and Test set and the relation between Spinach category and Labels.

| Category | Total Image | For Training Set | For Test Set | Target Label |
|---|---|---|---|---|
| Jute Spinach | 750 | 600 | 150 | 0 |
| Malabar Spinach | 758 | 606 | 152 | 1 |
| Red Spinach | 761 | 609 | 152 | 2 |
| Taro Spinach | 772 | 618 | 154 | 3 |
| Water Spinach | 744 | 595 | 149 | 4 |
|  | 3785 | 3028 | 757 |  |

In this section of the article, we will analyze all the results and compare them with each other to see which model is best for our classification and gives the best performance on our dataset.

Fig. 4. show the confusion matrix for the four CNN models. It is visualized the total correct and incorrect prediction results by the classifier. The predicted label is represented by the X-axis and the truth label is represented by Y-axis. A confusion matrix contains some number of True Positive(TP), True Negative(TN), False Positive(FP), and False Negative(FN). Suppose here when the classifier predicts it is jute spinach and the actual output is also jute spinach then it is True Positive(TP).

From Table 2 we observed, Malabar and Taro spinach have an accuracy of 99.74. This means that the InceptionV3 model can classify these two spinach better than others. Because of its False Positive Rate 0.33% but False Negative Rate 0%. Contrarily the accuracy of Water spinach has the lowest accuracy of 97.89% because of the total number of False Positive and False Negative 16.

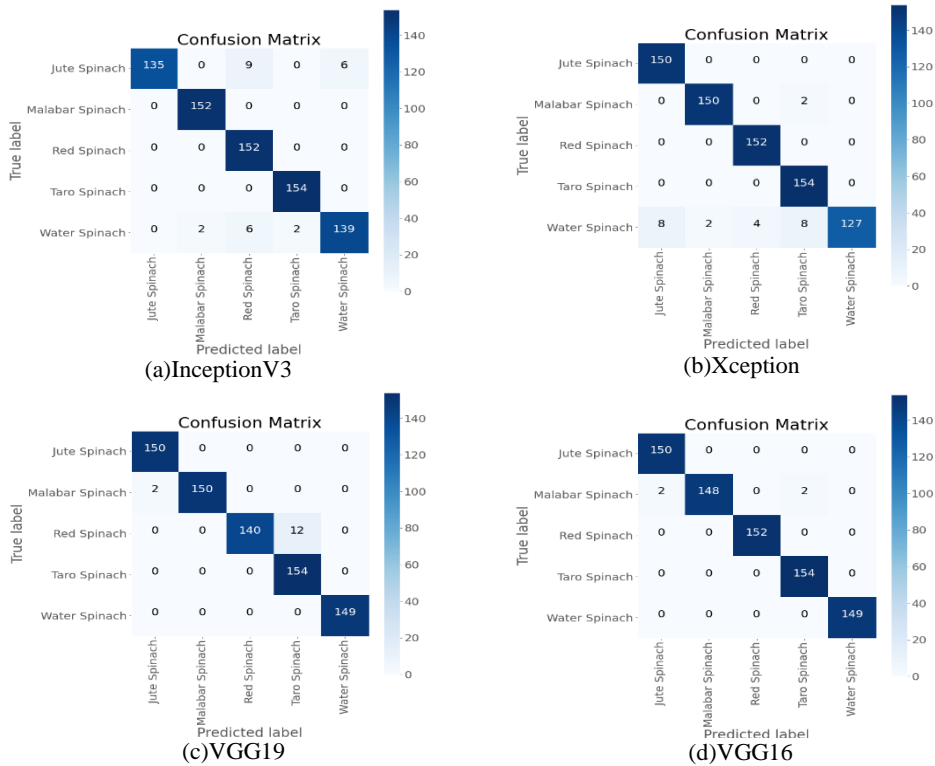

(a)InceptionV3      (b)Xception

(c)VGG19      (d)VGG16

**Fig. 4.** Confusion matrix for (a)InceptionV3, (b)Xception, (c)VGG19, (d)VGG16.

The obtained results by Xception (Table 3) tell us that its accuracy is higher in Malabar and Red spinach classification, which is 99.47%. The number of False Positive and False Negative in both of these is 4. As a result, the sum of False Positive Rate and False Negative Rate is equal. Like InceptionV3, it has the lowest accuracy of the Water spinach class which is 97.10%.

Concerning the VGG19 results (Table 4), we may notice that it achieved a 100% accuracy for the Water spinach class. Which is more than the rest of the classes. Here Jute spinach, Taro spinach, and Water spinach the Sensitivity of these 3 classes is 100%. With this, Precision and Specificity is 100% for Malabar spinach, Red spinach, and Water spinach. This model has low accuracy for the Red and Taro spinach class, which is 98.41%.

The results are given by VGG16 (Table 5) show that it achieves 100% accuracy for Red and Water spinach. And it has the lowest accuracy for Malabar spinach, which is 99.47%. For the Malabar spinach, Red spinach, and Water spinach class, we observe that it was identified with good specificity and precision which is 100%. And for Jute spinach, Red spinach, Taro spinach, Water spinach class Sensitivity is 100%.

**Table 2.** Evaluation metric for InceptionV3.

| Spinach Category | True Positive (TP) | True Negative (TN) | False Positive (FP) | False Negative (FN) | Precision (%) | F1 (%) | Sensitivity (%) | Specificity (%) | False Positive Rate (FPR) (%) | False Negative Rate (FNR) (%) | Accuracy (%) |
|---|---|---|---|---|---|---|---|---|---|---|---|
| Jute | 135 | 607 | 0 | 15 | 100 | 94.74 | 90.00 | 100 | 0 | 10 | 98.02 |
| Malabar | 152 | 603 | 2 | 0 | 98.70 | 99.35 | 100 | 99.67 | 0.33 | 0 | 99.74 |
| Red | 152 | 590 | 15 | 0 | 91.02 | 95.30 | 100 | 97.52 | 2.48 | 0 | 98.02 |
| Taro | 154 | 601 | 2 | 0 | 98.72 | 99.35 | 100 | 99.67 | 0.33 | 0 | 99.74 |
| Water | 139 | 602 | 6 | 10 | 95.86 | 94.56 | 93.29 | 99.01 | 0.99 | 6.71 | 97.89 |

The main performance summary of our experiment has been presented in Table 6. VGG16 has achieved the highest 99.79% accuracy. Out of the total 757 images of the test set, this classifier has been able to classify 753 images correctly. Only 4 images of Malabar spinach class have wrongly predicted, shown in (Fig. 8d). At the same time, it's Precision, F1 score, Sensitivity is 99.47% and Specificity is the highest at 99.87%. This table also tells us that InceptionV3 achieved the lowest accuracy in our experiment, which is 98.68%. In the case of a total of 25 images, the wrong prediction has been made by the InceptionV3 classifier. On the other hand, the Xception model made incorrect predictions for 24 images. As a result, it gave the second-lowest 98.73% accuracy. The difference in accuracy with the InceptionV3 model is only 0.5%. The VGG19 model gives the second-highest 99.26% accuracy. It has classified a total of 743 images correctly and incorrect classification for 14 images, shown in Table. 6. By analyzing all these results we can say that the VGG16 model Gives a best performance than the rest in our experiment.

**Table 3.** Evaluation metric for Xception.

| Spinach Category | True Positive (TP) | True Negative (TN) | False Positive (FP) | False Negative (FN) | Precision (%) | F1 (%) | Sensitivity (%) | Specificity (%) | False Positive Rate (FPR) (%) | False Negative Rate (FNR) (%) | Accuracy (%) |
|---|---|---|---|---|---|---|---|---|---|---|---|
| Jute | 150 | 599 | 8 | 0 | 94.94 | 97.40 | 100 | 98.68 | 1.32 | 0 | 98.94 |
| Malabar | 150 | 603 | 2 | 2 | 98.68 | 98.68 | 98.68 | 99.67 | 0.33 | 1.32 | 99.47 |
| Red | 152 | 601 | 4 | 0 | 97.44 | 98.70 | 100 | 99.34 | 0.66 | 0 | 99.47 |
| Taro | 154 | 593 | 10 | 0 | 93.90 | 96.86 | 100 | 98.34 | 1.66 | 0 | 98.68 |
| Water | 127 | 608 | 0 | 22 | 100 | 92.03 | 85.23 | 100 | 0 | 14.77 | 97.10 |

**Table 4.** Evaluation metric for VGG19.

| Spinach Category | True Positive (TP) | True Negative (TN) | False Positive (FP) | False Negative (FN) | Precision (%) | F1 (%) | Sensitivity (%) | Specificity (%) | False Positive Rate (FPR) (%) | False Negative Rate (FNR) (%) | Accuracy (%) |
|---|---|---|---|---|---|---|---|---|---|---|---|
| Jute | 150 | 605 | 2 | 0 | 98.68 | 99.34 | 100 | 99.67 | 0.33 | 0 | 99.74 |
| Malabar | 150 | 605 | 0 | 2 | 100 | 99.34 | 98.68 | 100 | 0 | 1.32 | 99.74 |
| Red | 140 | 605 | 0 | 12 | 100 | 95.89 | 92.11 | 100 | 0 | 7.89 | 98.41 |
| Taro | 154 | 591 | 12 | 0 | 92.77 | 96.25 | 100 | 98.01 | 1.99 | 0 | 98.41 |
| Water | 149 | 608 | 0 | 0 | 100 | 100 | 100 | 100 | 0 | 0 | 100 |

## 5    Conclusion & Future work

In this paper, we proposed a local spinach classification method using the CNN model, which can automatically classify the local spinach. Four classifiers proved image processing about similar-looking spinach can be classified & can give a high output through these classifiers. We trained our dataset with several deep learning based con-volutional neural networks. Then we used four different models to compare the accuracy between them. All of them gave high accuracy for our dataset. We trained those models with noiseless or few noise images so this model will be unable to give the best performance if the images have noise. As those models gives high performance, we expect that these results will help for further research. Moreover, it will help to develop an image application through the application, general people can classify the edible local spinach.

**Table 5.** Evaluation metric for VGG16.

| Spinach Category | True Positive (TP) | True Negative (TN) | False Positive (FP) | False Negative (FN) | Precision (%) | F1 (%) | Sensitivity (%) | Specificity (%) | False Positive Rate (FPR) (%) | False Negative Rate (FNR) (%) | Accuracy (%) |
|---|---|---|---|---|---|---|---|---|---|---|---|
| Jute | 150 | 605 | 2 | 0 | 98.68 | 99.34 | 100 | 99.67 | 0.33 | 0 | 99.74 |
| Malabar | 148 | 605 | 0 | 4 | 100 | 98.67 | 97.37 | 100 | 0 | 2.63 | 99.47 |
| Red | 152 | 605 | 0 | 0 | 100 | 100 | 100 | 100 | 0 | 0 | 100 |
| Taro | 154 | 601 | 2 | 0 | 98.72 | 99.35 | 100 | 99.67 | 0.33 | 0 | 99.74 |
| Water | 149 | 608 | 0 | 0 | 100 | 100 | 100 | 100 | 0 | 0 | 100 |

**Table 6.** Evaluation metric for Four CNN Models. InceptionV3, Xception, VGG19, VGG16.

| Models | True Positive (TP) | True Negative (TN) | False Positive (FP) | False Negative (FN) | Precision (%) | F1 (%) | Sensitivity (%) | Specificity (%) | False Positive Rate (FPR) (%) | False Negative Rate (FNR) (%) | Accuracy (%) |
|---|---|---|---|---|---|---|---|---|---|---|---|
| Inception3 | 732 | 3003 | 25 | 25 | 96.70 | 96.70 | 96.70 | 99.17 | 0.83 | 3.30 | 98.68 |
| Xception | 733 | 3004 | 24 | 24 | 96.83 | 96.83 | 96.83 | 99.21 | 0.79 | 3.17 | 98.73 |
| VGG19 | 743 | 3014 | 14 | 14 | 98.15 | 98.15 | 98.15 | 99.54 | 0.46 | 1.85 | 99.26 |
| VGG16 | 753 | 3024 | 4 | 4 | 99.47 | 99.47 | 99.47 | 99.87 | 0.13 | 0.53 | 99.79 |

## 6   Acknowledgments

We gratefully acknowledge for providing GPU support from Computational Intelligence Lab for providing the necessary support. We thank, Dept. of CSE, Daffodil Inte-rnational University. Moreover, thanks to the anonymous reviewers for their valuable comments and feedback.

**References**

1. Roughani, Mehdi, M., S.: Spinach: An important green leafy vegetable and medicinal herb. The 2nd International Conference on Medicinal Plants, Organic Farming, Natural and Pharmaceutical Ingredients(2019).
2. Islam, R., Ara, T.: Leafy Vegetables in Bangladesh. UBN: 015-A94510112010, Photon ebooks.(2015).
3. Wu, S., G., Bao, F., S., Xu, E., Y., Wang, Y., Chang , Y., Xiang, Q.: A Leaf Recognition Algorithm for Plant Classification Using Probabilistic Neural Network. 2007 IEEE International Symposium on Signal Processing and Information Technology, Giza, 2007, pp. 11-16, (2007).
4. Wang, G., Sun, Y., Wang, J.: Automatic Image-Based Plant Disease Severity Estimation Using Deep Learning. Computational Intelligence And Neuroscience, 2017, 1-8.(2017).
5. Sun, Y., Liu, Y., Wang, G., Zhang, H.: Deep Learning for Plant Identification in Natural Environment. Computational Intelligence And Neuroscience, 2017, 1-6. (2017).
6. Zhu, S., Feng, L., Zhang, C., Bao, Y., He, Y.: Identifying Freshness of Spinach Leaves Stored at Different Temperatures Using Hyperspectral Imaging. Foods 2019, 8, 356,(2019).
7. Yalcin, H., Razavi, S.: Plant classification using convolutional neural networks. 2016 Fifth International Conference on Agro-Geoinformatics (Agro-Geoinformatics), Tianjin, 2016, pp. 1-5,(2016).
8. Kaya, A., Keceli, A., Catal, C., Yalic, H., Temucin, H., Tekinerdogan, B.: Analysis of transfer learning for deep neural network based plant classification models. Computers And Electronics In Agriculture, 158, 20-29.(2019).
9. Priya, C., A., T., Balasaravanan, A., Thanamani, S.: An efficient leaf recognition algorithm for plant classification using support vector machine. International Conference on Pattern Recognition, Informatics and Medical Engineering (PRIME-2012), Salem, Tamilnadu, 2012, pp. 428-432, (2012).
10. Szegedy, C., Liu, W., Jia, Y., Sermanet, P., Reed, S., Anguelov, D.: Going deeper with convolutions. IEEE Conference on Computer Vision and Pattern Recognition (CVPR), p. 1–9, Boston, MA,(2015).
11. Szegedy, C., Vanhoucke, V., Ioffe, S., Shlens, J., Wojna, Z.: Rethinking the Inception Architecture for Computer Vision. 2016 IEEE Conference on Computer Vision and Pattern Recognition (CVPR), pp. 2818-2826, Las Vegas, NV, (2016).
12. Chollet, F.: Xception: deep learning with depthwise separable con- volutions. In: Proceedings of the IEEE conference on computer vision and pattern recognition. p. 1251–8 (2017).
13. Simonyan, K., Zisserman, A.: Very deep convolutional networks for large-scale image recognition. arXiv preprint arXiv :1409.1556, (2014).
14. Dalwinder, S., Birmohan, S.: Investigating the impact of data normalization on classification performance, Applied Soft Computing, 105524, ISSN 1568-4946,(2019).